\documentclass{article}

\usepackage{times}
\usepackage{amsthm}
\newtheorem{theorem}{Theorem}
\usepackage{graphicx} 
\usepackage{subfigure} 
\usepackage{amsmath}
\usepackage[]{units}

\usepackage{amsfonts}

\usepackage{algorithm}
\usepackage{algorithmic}

\usepackage[breaklinks=true]{hyperref}
\usepackage{breakcites}

\DeclareSymbolFont{symbolsC}{U}{pxsyc}{m}{n}
\DeclareMathSymbol{\coloneqq}{\mathrel}{symbolsC}{"42}

\newtheorem*{corollary}{Corollary}

\newtheorem*{definition}{Definition}

\usepackage[accepted]{icml2016}

\icmltitlerunning{Thresholding Bandit Problem}

\begin{document} 

\twocolumn[
\icmltitle{An optimal algorithm for the Thresholding Bandit Problem}

\icmlauthor{Andrea Locatelli}{andrea.locatelli@uni-potsdam.de}
\icmlauthor{Maurilio Gutzeit}{mgutzeit@uni-potsdam.de}
\icmlauthor{Alexandra Carpentier}{carpentier@uni-potsdam.de}
\icmladdress{Department of Mathematics, University of Potsdam, Germany}

\icmlkeywords{bandits, online learning, sequential, machine learning, ICML}

\vskip 0.3in
]

\begin{abstract} 
We study a specific \textit{combinatorial pure exploration stochastic bandit problem} where the learner aims at finding the set of arms whose means are above a given threshold, up to a given precision, and \textit{for a fixed time horizon}. We propose a parameter-free algorithm based on an original heuristic, and prove that it is optimal for this problem by deriving matching upper and lower bounds. To the best of our knowledge, this is the first non-trivial pure exploration setting with \textit{fixed budget} for which optimal strategies are constructed.
\end{abstract}

\section{Introduction}
\label{introduction}

In this paper we study a specific \textit{combinatorial, pure exploration, stochastic bandit setting}. More precisely, consider a  stochastic bandit setting where each arm has mean $\mu_k$. The learner can sample sequentially $T>0$ samples from the arms and aims at finding as efficiently as possible the set of arms whose means are larger than a threshold $\tau \in \mathbb R$. In this paper, we refer to this setting as the \textit{Thresholding Bandit Problem (TBP)}, which is a specific instance of the combinatorial pure exploration bandit setting introduced in~\cite{chen2014combinatorial}. A simpler "one armed" version of this problem is known as the SIGN-$\xi$ problem, see~\cite{chen2015optimal}.

This problem is related to the popular combinatorial pure exploration bandit problem known as the TopM problem where the aim of the learner is to return the set of $M$ arms with highest mean~\cite{bubeckmultiple,gabillon2012best, kaufmann2014complexity, zhou2014optimal,cao2015top} - which is a combinatorial version of the best arm identification problem~\cite{even2002pac,mannor2004sample, bubeck2009pure,audibert2010best, gabillon2012best, jamieson2013lil,karnin2013almost,kaufmann2014complexity,chen2015optimal}. To formulate this link with a simple metaphor, the TopM problem is a "contest" and the TBP problem is an "exam": in the former, the learner wants to select the $M$ arms with highest mean, in the latter the learner wants to select the arms whose means are higher than a certain threshold. We believe that this distinction is important and that in many applications the TBP problem is more relevant than the TopM, as in many domains one has a natural "efficiency", or "correctness" threshold above which one wants to use an option. For instance in industrial applications, one wants to keep a machine if its production's value is above its functioning costs, in crowd-sourcing one wants to hire a worker as long as its productivity is higher than its wage, etc. In addition to these applications derived from the TopM problem, the TBP problem has applications in dueling bandits and is a natural way to cast the problem of active and discrete level set detection, which is in turn related to the important applications of active classification, and active anomaly detection - we detail this point more in Subsection~\ref{sub:levelset}.

As mentioned previously, the TBP problem is a specific instance of the combinatorial pure exploration bandit framework introduced in~\cite{chen2014combinatorial}. Without going into the details of the combinatorial pure exploration setting for which the paper~\cite{chen2014combinatorial} derives interesting general results, we will summarize what these results imply for the particular TBP and TopM problems, which are specific cases of the combinatorial pure exploration setting. As it is often the case for pure exploration problems, the paper~\cite{chen2014combinatorial} distinguishes between two settings:
\begin{itemize}
\item The \textit{fixed budget setting} where the learner aims, given a fixed budget $T$, at returning the set of arms that are above the threshold (in the case of TBP) or the set of $M$ best arms (in the case of TopM), with highest possible probability. In this setting, upper and lower bounds are on the \textit{probability of making an error when returning the set of arms}.
\item The \textit{fixed confidence setting} where the learner aims, given a probability $\delta$ of acceptable error, at returning the set of arms that are above the threshold (in the case of TBP) or the set of $M$ best arms (in the case of TopM) with as few pulls of the arms as possible. In this setting, upper and lower bounds are on the \textit{number of pulls $T$ that are necessary to return the correct set of arm with probability at least $1-\delta$}.
\end{itemize}
The similarities and dissemblance of these two settings have been discussed in the literature in the case of the TopM problem (in particular in the case $M=1$), see~\cite{gabillon2012best, karnin2013almost,chen2014combinatorial}. While as explained in~\cite{audibert2010best, gabillon2012best}, the two settings share similarities in the specific case when additional information about the problem is available to the learner (such as the complexity $H$ defined in Table~\ref{f:tab}), they are very different in general and results do not transfer from one setting to the other, see~\cite{bubeck2009pure, audibert2010best, karnin2013almost,kaufmann2014complexity}. In particular we highlight the following fact: while the \textit{fixed confidence setting} is relatively well understood in the sense that there are constructions for optimal strategies~\cite{kalyanakrishnan2012pac, jamieson2013lil,karnin2013almost,kaufmann2014complexity, chen2015optimal}, there is an important knowledge gap in the \textit{fixed budget setting}. In this case, without the knowledge of additional information on the problem such as e.g.~the complexity $H$ defined in Table~\ref{f:tab}, there is a gap between the known upper and lower bounds, see~\cite{audibert2010best,gabillon2012best, karnin2013almost,kaufmann2014complexity}.
This knowledge gap is more acute for the general combinatorial exploration bandit problem defined in the paper~\cite{chen2014combinatorial} (see their Theorem 3) - and therefore for the TBP problem (where in fact no fixed budget lower bound exists to the best of our knowledge). 
We summarize in Table~\ref{f:tab} the state of the art results for the TBP problem and for the TopM problem with $M=1$.

\begin{table}[ht]
\renewcommand{\arraystretch}{1.4}
\begin{center}
\begin{tabular}{c||c|c}
  \hline
Problem  & Lower Bound & Upper Bound   \\
\hline
\hline
  TBP (FC) &  $H\log\big(\frac{1}{\delta}\big)$ & $H\log\big(\frac{1}{\delta}\big)$ \\
  TBP (FB) & No results & $K\exp\big(-\frac{T}{\log(K)H_2}\big)$\\
\hline      
    TopM (FC) &  $H\log\big(\frac{1}{\delta}\big)$ & $H\log\big(\frac{1}{\delta}\big)$ \\
    TopM (FB) & $\exp\big(-\frac{T}{H}\big)$& $K\exp\big(-\frac{T}{\log(K)H_2}\big)$\\

    \hline
\end{tabular}
\label{f:tab}
\end{center}
\caption{State of the art results for the TBP problem and the TopM problem with $M=1$ with fixed confidence $\Delta$ for $\delta$ small enough (FC) and fixed budget (FB) - for FC, bound on the expected total number of samples needed for making an error of at most $\delta$ on the set of arms and for FB, bound on the probability of making a mistake on the returned set of arms. The quantities $H, H_2$ depend on the means $\mu_k$ of the arm distributions  and are defined in~\cite{chen2014combinatorial} (and are not the same for TopM and TBP). In the case of the TBP problem, set $\Delta_k = |\tau - \mu_k|$ and set $\Delta_{(k)}$ for the $\Delta_k$ ordered in increasing order, we have $H = \sum_i \Delta_i^{-2}$ and $H_2 = \min_i i \Delta_{(i)}^{-2}$. For the TopM problem with $M=1$, the same definitions holds with $\Delta_k = \max_i \mu_i - \mu_k$.}
\vspace{-0.8cm}
\end{table}


The summary of Table~\ref{f:tab} highlights that in the fixed budget setting, both for the TopM and the TBP problem, the correct \textit{complexity} $H^*$ that should appear in the bound, i.e.~what is the problem dependent quantity $H^*$ such that the upper and lower bounds on the probability of error  is of order $\exp(-n/H^*)$, is still an open question. In the TopM problem, Table~\ref{f:tab} implies that $H \leq H^* \leq \log(2K) H_2$. In the TBP problem, Table~\ref{f:tab} implies $0 \leq H^* \leq \log(2K) H_2$, since to the best of our knowledge a lower bound for this problem exists only in the case of the fixed confidence setting. Note that although this gap may appear small in particular in the case of the TopM problem as it involves "only" a $\log(K)$ multiplicative factor, it is far from being negligible since the $\log(K)$ gap factor acts on a term of order exponential minus $T$ exponentially.

In this paper we close, up to constants, the gap in the fixed budget setting for the TBP problem - we prove that $H^*=H$. In addition, we also prove that our strategy minimizes at the same time the cumulative regret, and identifies optimally the best arm, provided that the highest mean of the arms is known to the learner. Our findings are summarized in Table~\ref{f:tab2}. In order to do that, we introduce a new algorithm for the TBP problem which is entirely \textit{parameter free}, and based on an original heuristic. In Section~\ref{TBP}, we describe formally the TBP problem, the algorithm, and the results. In Section~\ref{extensions}, we describe how our algorithm can be applied to the active detection of discrete level sets, and therefore to the problem of active classification and active anomaly detection. We also describe what are the implications of our results for the TopM problem. Finally Section~\ref{experiments} presents some simulations for evaluating our algorithm with respect to the state of the art competitors. The proofs of all theorems are in Appendix \ref{app:A}, as well as additional simulation results.
\begin{table}[H]
\centering
\renewcommand{\arraystretch}{1.3}
\begin{center}
\begin{tabular}{c||c}
  \hline
Problem  & Results   \\
\hline
\hline
    TBP (FB) : UB & $\exp\big(-\frac{T}{H} + \log\big(\log(T)K\big)\big)$\\
    TBP (FB) : LB & $\exp\big(-\frac{T}{H} - \log\big(\log(T)K\big)\big)$\\
\hline      
    TopM (FB) : UB & $\exp\big(-\frac{T}{H} + \log\big(\log(T)K\big)\big)$\\
    \footnotesize{(with $\mu^*$ known)} & \quad \\

    \hline
\end{tabular}
\label{f:tab2}
\end{center}
\caption{Our results for the TopM and the TBP problem in the fixed budget setting - i.e.~upper and lower bounds on the probability of making a mistake on the set of arms returned by the learner.}
\end{table}

\section{The Thresholding Bandit Problem}\label{TBP}
\subsection{Problem formulation}
\label{setting}
{\bf Learning setting} Let $K$ be the number of arms that the learner can choose from. Each of these arms is characterized by a distribution $\nu_k$ that we assume to be R-sub-Gaussian.
\begin{definition}[$R$-sub-Gaussian distribution]\label{assum:sg}
Let $R>0$. A distribution $\nu$ is $R$-sub-Gaussian if for all $t \in \mathbb{R}$ we have 
\vspace{-0.1cm}
\begin{equation*}
\mathbb{E}_{X \sim \nu}[\exp(tX - t\mathbb{E}[X])] \leq \exp(R^2t^2/2).
\end{equation*}
\vspace{-0.8cm}
\end{definition}
This encompasses various distributions such as bounded distributions or Gaussian distributions of variance $R^2$ for $R \in \mathbb R$. Such distributions have a finite mean, let $\mu_k = \mathbb{E}_{X \sim \nu_k}[X]$ be the mean of arm $k$.

We consider the following dynamic game setting which is common in the bandit literature. For any time $t \geq 1$, the learner chooses an arm $I_t$ from $\mathbb{A} = \{1,...,K\}$. It receives a noisy reward drawn from the distribution $\nu_{I_t}$ associated to the chosen arm. An adaptive learner bases its decision at time $t$ on the samples observed in the past.\

{\bf Set notations} Let $u \in \mathbb{R}$ and $\mathbb{A}$ be the finite set of arms. We define $S_u$ as the set of arms whose means are over $u$, that is $S_u:=\{k \in \mathbb{A}, \mu_k \geq u\}$. We also define $S_u^C$ as the complimentary set of $S_u$ in $\mathbb{A}$, i.e. $S_u^C=\{k \in \mathbb{A}, \mu_k < u\}$.

{\bf Objective} Let $T>0$ (not necessarily known to the learner beforehand) be the horizon of the game, let $\tau \in \mathbb{R}$ be the {\it threshold} and $\epsilon \geq 0$ be the {\it precision}. We define the $(\tau, \epsilon)$ thresholding problem as such : after $T$ rounds of the game described above, the goal of the learner is to correctly identify the \textit{arms whose means are over or under the threshold $\tau$ up to a certain precision $\epsilon$}, i.e.~to correctly discriminate arms that belong to $S_{\tau + \epsilon}$ from those in $S_{\tau-\epsilon}^C$. In the rest of the paper, the sentence "the arm is over the threshold $\tau$" is to be understood as "the arm's mean is over the threshold".

After $T$ rounds of the previously defined game, the learner has to output a set $\widehat{S}_\tau := \widehat{S}_\tau(T) \subset \mathbb{A}$ of arms and it suffers the following loss:
$$\mathcal{L}(T) = \mathbb{I}(S_{\tau + \epsilon} \cap \widehat{S}_\tau^C \not = \emptyset \quad \lor \quad S_{\tau - \epsilon}^C \cap \widehat{S}_\tau \not = \emptyset).$$
A good learner minimizes this loss by correctly discriminating arms that are outside of a $2\epsilon$ band around the threshold: arms whose means are smaller than $(\tau - \epsilon)$ should not belong to the output set $\widehat{S}_\tau$, and symmetrically those whose means are bigger than $(\tau + \epsilon)$ should not belong to $\widehat{S}_\tau^C$. If it manages to do so, the algorithm suffers no loss and otherwise it incurs a loss of $1$. For arms that lie inside this $2\epsilon$ strip, mistakes on the other hand bear no cost. If we set $\epsilon$ to $0$ we recover the exact TBP thresholding problem described in the introduction, and the algorithm suffers no loss if it discriminates exactly arms that are over the threshold from those under.

Let $\mathbb{E}$ be the expectation according to the samples collected by an algorithm, its expected loss is:
$$\mathbb{E}[\mathcal{L}(T)] = \mathbb{P}(S_{\tau + \epsilon} \cap \widehat{S}_\tau^C \not = \emptyset \quad \lor \quad S_{\tau - \epsilon}^C \cap \widehat{S}_\tau \not = \emptyset),$$ 
i.e.~it is the probability of making a mistake, that is rejecting an arm over $(\tau + \epsilon)$ or accepting an arm under $(\tau - \epsilon)$. The lower this probability of error, the better the algorithm, as an oracle strategy would simply rightly classify each arm and suffer an expected loss of $0$.

Our problem is a pure exploration bandit problem, and is in fact, shifting the means by $-\tau$, a specific case of the pure exploration bandit problem considered in~\cite{chen2014combinatorial} - namely the specific case where the set of sets of arms that they call $\mathcal M$ and which is their decision class is the set of all possible set of arms. We will comment more on this later in Subsection~\ref{ss:disc}.

{\bf Problem complexity} We define $\Delta_i^{\tau,\epsilon}$ the gap of arm $i$ with respect to $\tau$ and $\epsilon$  as:
\vspace{-0.3cm}
\begin{equation}\label{def_delta}
\Delta_i := \Delta_i^{\tau,\epsilon} = |\mu_i - \tau| + \epsilon.
\end{equation}
We also define the complexity $H_\epsilon$ of the problem as
\vspace{-0.3cm}
\begin{equation}\label{eq:H}
H := H_{\tau,\epsilon} = \sum_{i=1}^K (\Delta_i^{\tau,\epsilon})^{-2}.
\vspace{-0.2cm}
\end{equation}
We call $H$ complexity as it is a characterization of the hardness of the problem. A similar quantity was introduce for general combinatorial bandit problems~\cite{chen2014combinatorial} and is similar in essence to the complexity introduced for the best arm identification problem, see~\cite{audibert2010best}.




\subsection{A lower bound}\label{sub:LB}
\label{lowerbound}
In this section, we exhibit a lower bound for the thresholding problem. More precisely, for any sequence of gaps $(d_k)_k$, we define a finite set of problems where the distributions of the arms of these problems correspond to these gaps and are Gaussian of variance $1$. We lower bound the largest probability of error among these problems, for the best possible algorithm. 

\begin{theorem}\label{thm:LB}
Let $K,T \geq 0$. Let for any $i\leq K$, $d_i \geq 0$. Let $\tau \in \mathbb R, \epsilon > 0$.

For $0\leq i\leq K$, we write $\mathcal B^i$ for the problem where the distribution of arm $j\in \{1, \ldots, K\}$ is $\mathcal N(\tau+d_i+\epsilon,1)$ if $i \neq j$ and $\mathcal N(\tau-d_i-\epsilon,1)$ otherwise. For all these problems, $H:=H_{\tau, \epsilon} = \sum_i (d_i +2 \epsilon)^{-2}$ is the same by definition.

It holds that for any bandit algorithm
\vspace{-0.3cm}
\begin{align*}
\max\limits_{i \in \{0,\ldots,K\}}&  \mathbb{E}_{\mathcal{B}^i}(\mathcal L(T)) \geq \exp\big(-3T/H-\\
&4\log(12(\log(T)+1)K)\big),
\vspace{-0.2cm}
\end{align*}
where $\mathbb E_{\mathcal B^i}$ is the expectation according to the samples of problem $\mathcal B^i$.
\end{theorem}

This lower bound implies that even if the learner is given the distance of the mean of each arm to the threshold and the shape of the distribution of each arm (here Gaussian of variance $1$), any algorithm still makes an error of at least $\exp(-3T/H - 4\log(12(\log(T)+1)K))$ on one of the problems. This is a lower bound in a very strong sense because we really restrict the set of possibilities to a setting where we know all gaps and prove that nevertheless this lower bounds holds. Also it is non-asymptotic and holds for any $T$, and implies therefore a non-asymptotic minimax lower bound. The closer the means of the distributions to the threshold, the larger the complexity $H$, and the larger the lower bound. The proof is to be found in Appendix \ref{app:A}. 

This theorem's lower bound contains two terms in the exponential, a term that is linear in $T$ and a term that is of order $\log((\log(T)+1)K) \approx \log(\log(T)) + \log(K)$. For large enough values of $T$, one has the following simpler corollary.
\begin{corollary}\label{cor:LB}
Let $\bar H>0$ and $R>0$, $\tau \in \mathbb R$ and $\epsilon\geq 0$. Consider $\mathbb B_{\bar H,R}$ the set of $K$-armed bandit problems where the distributions of the arms are $R$-sub-Gaussian and which have all a complexity smaller than $\bar H$.

Assume that $T \geq 4\bar HR^2\log(12(\log(T)+1)K)$. It holds that for any bandit algorithm
\begin{align*}
\sup\limits_{\mathcal B \in \mathbb B_{\bar H, R}}&  \mathbb{E}_{\mathcal{B}}(\mathcal L(T)) \geq \exp\big(-4T/(R^2\bar H)\big),
\end{align*}
where $\mathbb E_{\mathcal B}$ is the expectation according to the samples of problem $\mathcal B \in \mathbb B_{\bar H, R}$.
\end{corollary}

\subsection{Algorithm APT and associated upper bound}
\label{algo}

In this section we introduce APT (Anytime Parameter-free Thresholding algorithm), an anytime parameter-free learning algorithm. Its heuristic is based on a simple observation, namely that a near optimal static strategy that  allocates $T_k$ samples to arm $k$ is such that $T_k \Delta_k^2$ is constant across $k$ (and increasing with $T$) - see Theorem~\ref{thm:LB}, 
and in particular the second half of Step 3 of its proof in Appendix~\ref{app:A} - and that therefore a natural idea is to simply pull at time $t$ the arm that minimizes an estimator of this quantity. Note that in this paper, we consider for the sake of simplicity that each arm is tested against the same threshold, however this can be relaxed to $(\tau_k)_k$ at no additional cost.

{\bf Algorithm} The algorithm receives as input the definition of the problem $(\tau, \epsilon)$. First, it pulls each arm of the game once. At time $t > K$, APT updates $T_i(t)$, the number of pulls up to time $t$ of arm $i$, and the empirical mean $\hat{\mu}_i(t)$ of arm $k$ after $T_i(t)$ pulls. Formally, for each $k \in \mathbb{A}$ it computes 
$T_i(t) = \sum_{s=1}^{t} \mathbb{I}(I_s = i)$ 
and the updated means
\begin{equation}\label{emp_mean}
\widehat{\mu}_i(t)= \frac{1}{T_i(t)} \sum_{s=1}^{T_i(t)} X_{i,s},
\end{equation}
where $X_{i,s}$ denotes the sample received when pulling $i$ for the $s$-th time. The algorithm then computes:
\begin{equation}\label{gap_k_t}
\widehat{\Delta}_i(s) := \widehat{\Delta}_i^{\tau,\epsilon}(s) = |\hat{\mu}_i(t) - \tau| + \epsilon,
\end{equation}
the current empirical estimate of the gap associated with arm $i$. The algorithm then computes:
\begin{equation}\label{b_k_t}
B_i(t+1) = \sqrt[]{T_i(t)}\widehat{\Delta}_i(t).
\end{equation}
and pulls the arm $I_{t+1} = \arg\min\limits_{i \leq K} B_i(t+1)$ that minimizes this quantity.
At the end of the horizon $T$, the algorithm outputs the set of arms $\widehat{S}_\tau = \{k: \widehat{\mu}_k(T) \geq \tau\}.$

\begin{algorithm}[tb]
   \caption{APT algorithm}
   \label{alg:example}
\begin{algorithmic}
   \STATE {\bfseries Input:} $\tau$, $\epsilon$
   \STATE Pull each arm once 
   \FOR{$t=K+1$ {\bfseries to} $T$}
   \STATE Pull arm $I_t = \arg\min\limits_{k \leq K} B_k(t)$ from Equation~\eqref{b_k_t}
   \STATE Observe reward $X \sim \nu_{I_t}$
   \ENDFOR
   \STATE {\bfseries Output:} $\hat{S}_\tau = \{k: \hat{\mu}_k(T) \geq \tau\}$
\end{algorithmic}
\end{algorithm}



The expected loss of this algorithm can be bounded as follows.
\begin{theorem}\label{thm:UB} Let $K \geq 0, T \geq 2K$, and consider a problem $\mathcal B$. Assume that all arms $\nu_k$ of the problem are $R$-sub-Gaussian with means $\mu_k$. Let $\tau \in \mathbb R, \epsilon \geq 0$


Algorithm APT's expected loss is upper bounded on this problem as
\vspace{-0.2cm}
$$\mathbb E(\mathcal L(T)) \leq \exp\left(-\frac{1}{64R^2}\frac{T}{ H} + 2\log((\log(T)+1)K)\right),$$
where we remind that $ H = \sum_i (|\mu_i - \tau|+\epsilon)^{-2}$ and where $\mathbb E$ is the expectation according to the samples of the problem.
\end{theorem}

The bound of Theorem~\ref{thm:UB} holds for any $R$-sub-Gaussian bandit problem. \textit{Note that one does not need to know $R$ in order to implement the algorithm}, e.g.~ if the distributions are bounded, one does not need to know the bound. This is a desirable feature for an algorithm, yet e.g.~all algorithms based on upper confidence bounds need a bound on $R$. This bound is non-asymptotic (one just needs $T \geq 2K$ so that one can initialize the algorithm) and therefore Theorem~\ref{thm:UB} provides a minimax upper bound result over the class of problems that have sub-Gaussian constant $R$ and complexity $H$.

The term in the exponential of the lower bound of Theorem~\ref{thm:UB} matches the lower bound of Theorem~\ref{thm:LB} up to a multiplicative factor and the $\log((\log(T)+1)K)$ term. Now as in the case of the lower bound, for large enough values of $T$, one has the following simpler corollary.

\begin{corollary}\label{cor:UB}
Let $\bar H>0$ and $R>0$, $\tau \in \mathbb R$ and $\epsilon\geq 0$. Consider $\mathbb B_{\bar H,R}$ the set of $K$-armed bandit problems where the distributions of the arms are $R$-sub-Gaussian and whose complexity is smaller than $\bar H$.

Assume that $T \geq 256 \bar HR^2\log((\log(T)+1)K)$. For Algorithm APT it holds that
\vspace{-0.2cm}
\begin{align*}
\sup\limits_{\mathcal B \in \mathbb B_{\bar H, R}}&  \mathbb{E}_{\mathcal{B}}(\mathcal L(T)) \leq \exp\big(-T/(128R^2H)\big),
\vspace{-0.2cm}
\end{align*}
where $\mathbb E_{\mathcal B}$ is the expectation according to the samples of problem $\mathcal B \in \mathbb B_{\bar H, R}$
\end{corollary}
This corollary and Corollary~\ref{cor:LB} imply that for $T$ large enough - i.e.~of larger order than $HR^2\log((\log(T)+1)K)$ - Algorithm APT is order optimal over the class of problems whose complexity is bounded by $\bar H$ and whose arms are $R$-sub-Gaussian.

\subsection{Discussion}\label{ss:disc}

\paragraph{A parameter free algorithm} An important point that we want to highlight for our strategy APT is that it does not need any parameter, such as the complexity $H$, the horizon $T$ or the sub-Gaussian constant $R$. This contrasts with any upper confidence based approach as in e.g.~\cite{audibert2010best, gabillon2012best} (e.g.~the UCB-E algorithm in~\cite{audibert2010best}), which need as parameter an upper bound on $R$ and the exact knowledge of $H$, while the bound of Theorem~\ref{thm:UB} will hold for any $R$ and any $H$, and our algorithm adapts to these quantities. Also we would like to highlight that for the related problem of best arm identification, existing fixed budget strategies need to know the budget $T$ in advance~\cite{audibert2010best, karnin2013almost,chen2014combinatorial} - while our algorithm can be stopped at any time and the bound of Theorem~\ref{thm:UB} will hold.

\paragraph{Extensions to distributions that are not sub-Gaussian as opposed to adaptation to sub- models} It is easy to see in the light of~\cite{bubeck2013bandits} that one could extend our algorithm to non sub-Gaussian distributions by using an estimator other than the empirical means, as e.g.~the estimators in ~\cite{catoni2012challenging} or in~\cite{alon1996space}. These estimators have sub-Gaussian concentration asymptotically under the only assumption that the distributions have a finite $(1+v)$ moment with $v >0$ (and the sub-Gaussian concentration will depend on $v$). Using our algorithm with a such estimator will therefore provide a result that is similar to the one of Theorem~\ref{thm:UB} - and that without requiring the knowledge of $v$, which means that our algorithm APT modified for using these robust estimators instead of the empirical mean will work \textit{for any bandit problem where the arm distributions have a finite $(1+v)$ moment with $v>0$}.\\
On the other hand, if we consider more specific, e.g. exponential, models, it is possible to obtain a refined lower bound in terms of Kullback- Leibler divergences rather than gaps following ~\cite{kaufmann2014complexity}. However, an upper bound of the same order clearly comes at the cost of a more complicated strategy and holds in less generality than our bound.

\paragraph{Optimality of our strategy} As explained previously, the upper bound on the expected risk of algorithm APT is comparable to the lower bound on the expected risk up to a $\log\big((\log(T)+1)K\big)$ term (see Theorems~\ref{thm:UB} and Theorems~\ref{thm:LB}) - and this term vanishes when the horizon $T$ is large enough, namely when $T \geq O(HR^2 \log\big((\log(T)+1)  K\big))$, which is the case for most problems. So for $T$ large enough, our strategy is order optimal over the class of problems that have complexity smaller than $H$ and sub-Gaussian constant smaller than $R$.

\paragraph{Comparison with existing results} Our setting is a specific combinatorial pure exploration setting with fixed budget where the objective is to find the set of arms that are above a given threshold. Settings related to ours have been analyzed in the literature and the state of the art result on our problem can be found (to the best of our knowledge) in the paper~\cite{chen2014combinatorial}. In this paper, the authors consider a general pure exploration combinatorial problem. Given a set $\mathcal M$ of subsets of $\{1, \ldots, K\}$, they aim at finding a subset of arms $M^*\in \mathcal M$ such that $M^* = \arg\max_{M \in \mathcal M} \sum_{k \in M^*} \mu_k$. In the specific case where $\mathcal M$ is the set of all subsets of $\{1, \ldots, K\}$, their problem in the \textit{fixed budget setting} is exactly the same as ours when $\epsilon =0$ and the means are shifted by $-\tau$. Their algorithm CSAR's upper bound on the loss is (see their Theorem 3):
\vspace{-0.3cm}
\begin{equation*}
\mathbb E(\mathcal L(T)) \leq K^2\exp\Big(-\frac{T-K}{72R^2\log(K)H_{\text{CSAR},2}}\Big),
\vspace{-0.2cm}
\end{equation*}
where $H_{\text{CSAR},2} = \max_i i\Delta_{(i)}^{-2}$. As $H_{\text{CSAR},2}\log(K) \geq H$ by definition, there is a gap for their strategy in the fixed budget setting with respect to the lower bound of Theorem~\ref{thm:LB}, which is smaller and of order $\exp(-T/(HR^2))$. Our strategy on the contrary does not have this gap, and improves over the CSAR strategy. We believe that this lack of optimality for CSAR is not an artefact of the proof of the paper~\cite{chen2014combinatorial}, and that CSAR is sub-optimal, as it is a successive reject algorithm with fixed and non-adaptive reject phase length. A similar gap between upper and lower bounds for successive reject based algorithms in the \textit{fixed budget setting} was also observed for the best arm identification problem when no additional information such as the complexity are known to the learner, see~\cite{audibert2010best,karnin2013almost,kaufmann2014complexity,chen2014combinatorial}. It is therefore an interesting fact that there is a \textit{parameter free} optimal algorithm for our \textit{fixed budget} problem.

The paper~\cite{chen2014combinatorial} also provides results in the \textit{fixed confidence setting}, where the objective is to provide an $\epsilon$ optimal set using the smallest possible sample size. In these results such a gap in optimality does not appear and the algorithm CLUCB they propose is almost optimal, see also~\cite{kalyanakrishnan2012pac, jamieson2013lil,karnin2013almost,kaufmann2014complexity, chen2015optimal} for related results in the fixed confidence setting. This highlights that the fixed budget setting and the fixed confidence setting are fundamentally different (at least in the absence of additional information such as the complexity $H$), and that providing optimal strategies in the fixed budget setting is a more difficult problem than providing an adaptive strategy in the fixed confidence problem - 
adaptive algorithms that are nearly optimal in the absence of additional information have only been exhibited in the latter case. To the best of our knowledge, all strategies except ours have such an optimality gap for fixed budget pure exploration combinatorial bandit problems, while there exists fixed confidence strategies for general pure exploration combinatorial bandits that are very close to optimal, see~\cite{chen2014combinatorial}.

Now in the case where the learner has additional information on the problem, as e.g.~the complexity $H$, it has been proved in the TopM problem that a UCB-type strategy  has probability of error upper bounded as $\exp(-T/H)$, see~\cite{audibert2010best,gabillon2012best}. A similar UCB type of algorithm would also work in the TBP problem, implying the same upper bound results as APT. But we would like to highlight that the \textit{exact} knowledge of $H$ is needed by these algorithms for reaching this bound - which is unlikely in applications. Our strategy on the other hand reaches, up to constants, the optimal expected loss for the TBP problem, without needing any parameter.

\section{Extensions of our results to related settings}
\label{extensions}

In this section we detail some implications of the results of the previous section to some specific problems.

\subsection{Active level set detection : Active classification and active anomaly detection}\label{sub:levelset}

Here we explain how a simple modification of our setting transforms it into the setting of \textit{active level set detection}, and therefore why it can be applied to active classification and active anomaly detection. We define the problem of discrete, active level set detection as the problem of deciding as efficiently as possible, in our bandit setting, whether for any $k$ the probabilities that the samples of arms $\nu_k$ are above or below a given level $L$ are higher or smaller than a threshold $\tau$ up to a precision $\epsilon$, i.e.~it is the problem of deciding for all $k$ whether
$\tilde \mu_k(L):=\mathbb P_{X \sim\nu_k} (X >L) \geq \tau$, or not up to a precision $\epsilon$.

This problem can be immediately solved by our approach with a simple change of variable. Namely, for the sample $X_t \sim \nu_{I_t}$ collected by the algorithm at time $t$, consider the transformation $\tilde X_t = \mathbf 1_{X_t >L}$.
Then $\tilde X_t$ is a Bernoulli random variable of parameter $\tilde \mu_{I_t}(L)$ (which is a $1/2$-sub-Gaussian distribution) - and applying our algorithm to the transformed samples $\tilde X_t$ solves the active level set detection problem. This has two interesting applications, namely in active binary classification and in active anomaly detection.

\paragraph{Active binary classification}
In active binary classification, the learner aims at deciding, for $k$ points (the arms of the bandit), whether each point belongs to the class $1$ or the class $0$.

At each round $t$, the learner can request help from a homogeneous mass of experts (which can be a set of previously trained classifiers, where one wants to minimize the computational cost, or crowd-sourcing, when one wants to minimize the costs of the task), and obtain a {\it noisy} label for the chosen data point $I_t$. We assume that for any point $k$, the expert's responses are independent and stochastic random variables in $\{0,1\}$ of mean $\mu_k$ (i.e.~the arm distributions are Bernoulli random variables of parameter $\mu_k$). We assume that the experts are right on average and that the label $l_k$ of $k$ is equal to
$l_k := \mathbf 1\{\tilde\mu_k >1/2\}$.
The active classification task therefore amounts to deciding whether $\mu_k > \tau :=1/2$ or not, possibly up to a given precision $\epsilon$. Our strategy therefore directly applies to this problem by choosing $\tau = 1/2$.

\paragraph{Active anomaly detection}

In the case anomaly detection, a common way to characterize anomalies is to describe them as naturally {\it not concentrated} \cite{steinwart2005classification}. A natural way to characterize anomalies is thus to define a {\it cutoff level $L$}, and classify the samples e.g.~above this level $L$ as anomalous. Such an approach has already received attention for anomaly detection e.g  in~\cite{streeter2006selecting}, albeit in a cumulative regret setting.

Here we consider an active anomaly detection setting where we face $K$ sources of data (the arms), and we aim at sampling them actively to detect which sources emit anomalous samples with a probability higher than a given threshold $\tau$ - this threshold is chosen e.g.~as the maximal tolerable amount of anomalous behavior of a source. This illustrates the fact that as described in \cite{steinwart2005classification}, the problem of anomaly detection is indeed a problem of level set detection - and so the problem of active anomaly detection is a problem of active level set detection on which we can use our approach as explained above.

\subsection{Best arm identification and cumulative reward maximization with known highest mean value}
Two classical bandit problems are the best-arm identification problem and the cumulative reward maximization problem. In the former, the goal of the learner is to identify the arm with the highest mean~\cite{bubeck2009pure}. In the latter, the goal is to maximize the sum of the samples collected by the algorithm up to time $T$~\cite{auer1995gambling}. Intuitively, both problems should call for different strategies - in the best arm identification problem one wants to explore all arms heavily while in the cumulative reward maximization problem one wants to sample as much as possible the arm with the highest mean. Such intuition is backed up by Theorem 1 of \cite{bubeck2009pure}, which states that in the absence of additional information and with a fixed budget, the lower the regret suffered in the cumulative setting, expressed in terms of rewards, the higher the regret suffered in the identification problem, expressed in terms of probability of error. We prove in this section the somewhat non intuitive fact that if one knows the value of best arm's mean, its possible to perform both tasks \textit{simultaneously} by running our algorithm where we choose $\epsilon =0$ and $\tau = \mu^* := \max_k \mu_k$. Our algorithm then reduces to the $GCL^*$ algorithm that can be found in~\cite{salomon2011deviations}.

\paragraph{Best arm identification} In the best arm identification problem, the game setting is the same as the one we considered but the goal of the learner is different: it aims at returning an arm $J_T$ that with the highest possible mean. The following proposition holds for our strategy APT that runs for $T$ times, and then returns the arm $J_T$ that was the most pulled.
\begin{theorem}\label{thm:Simple}
Let $K>0$, $R >0$ and $T \geq 2K$ and consider a problem where the distribution of the arms $\nu_k$ is $R$-sub-Gaussian and has mean $\mu_k$. Let $\mu^* := \max_k \mu_k$ and $H_{\mu^*} = \sum_{i:\mu_i \neq \mu^*} (\mu^* - \mu_i)^{-2}$.

Then APT run with parameters $\tau = \mu^*$ and $\epsilon = 0$, recommending the arm $J_T = \arg\max\limits_{k \in \mathbb{A}} T_k(T)$, is such that
\vspace{-0.3cm}
\begin{equation*}
\mathbb{P}(\mu_{J_T} \neq \mu^*) \leq \exp\big(-\frac{T}{36R^2H_{\mu^*}} + 2\log(\log(T)+1)K\big).
\end{equation*}
\vspace{-0.8cm}
\end{theorem}
If the complexity $H$ is also known to the learner, algorithm UCB-E from~\cite{audibert2010best} would attain a similar performance.

\begin{figure*}[ht]\label{Figure1}
\centering
\includegraphics[scale=0.3]{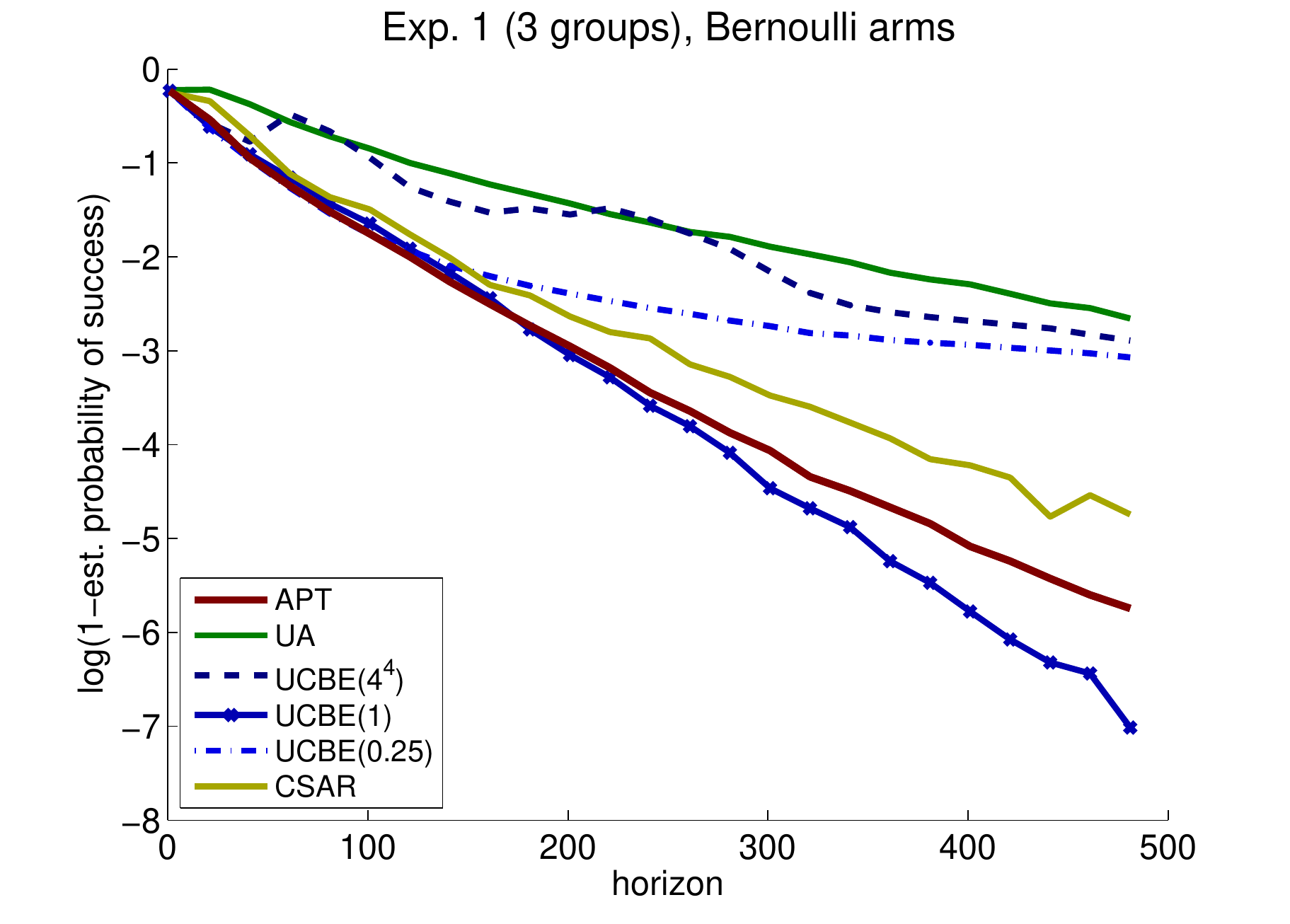}\ 
\includegraphics[scale=0.3]{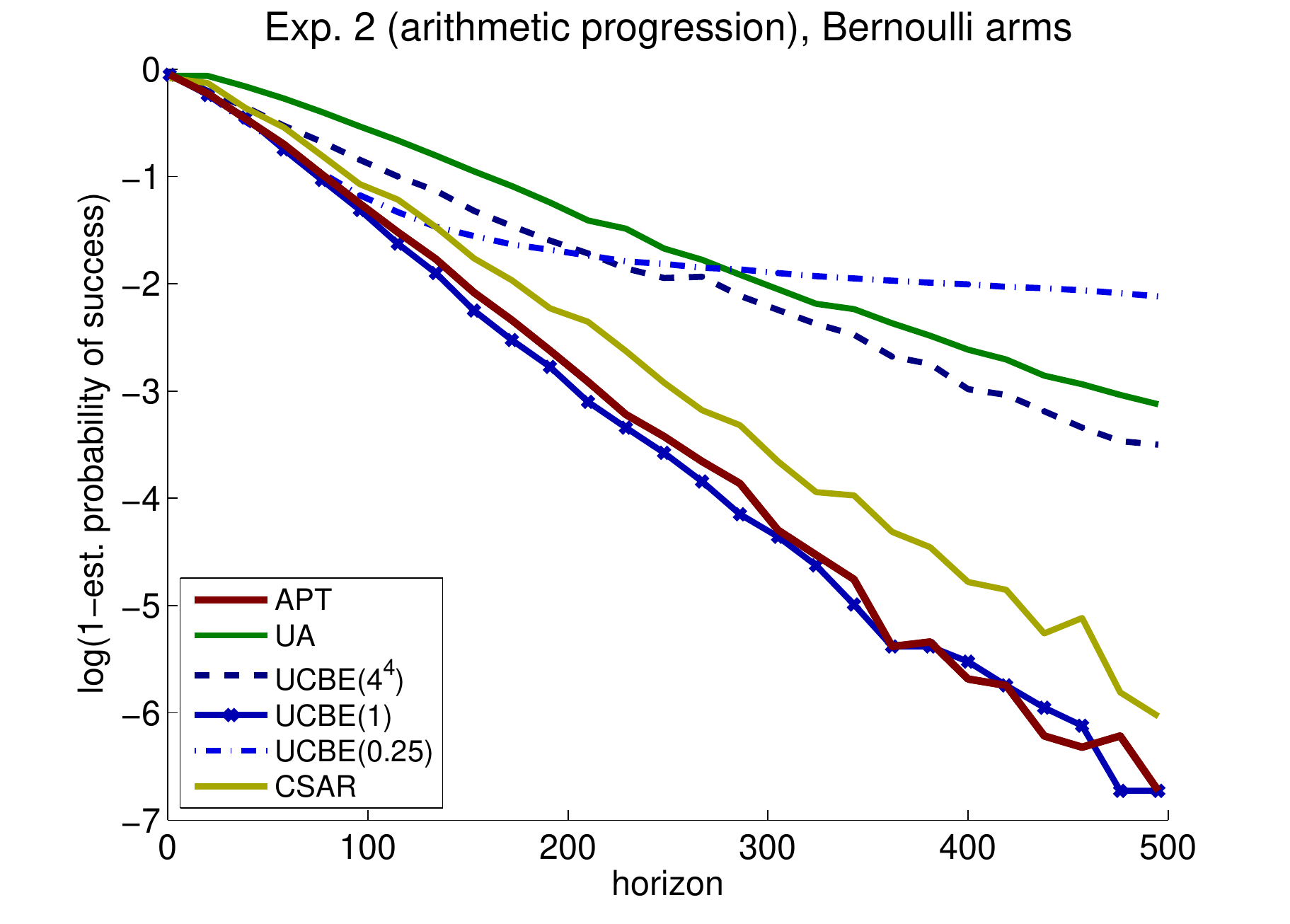}\ 
\includegraphics[scale=0.3]{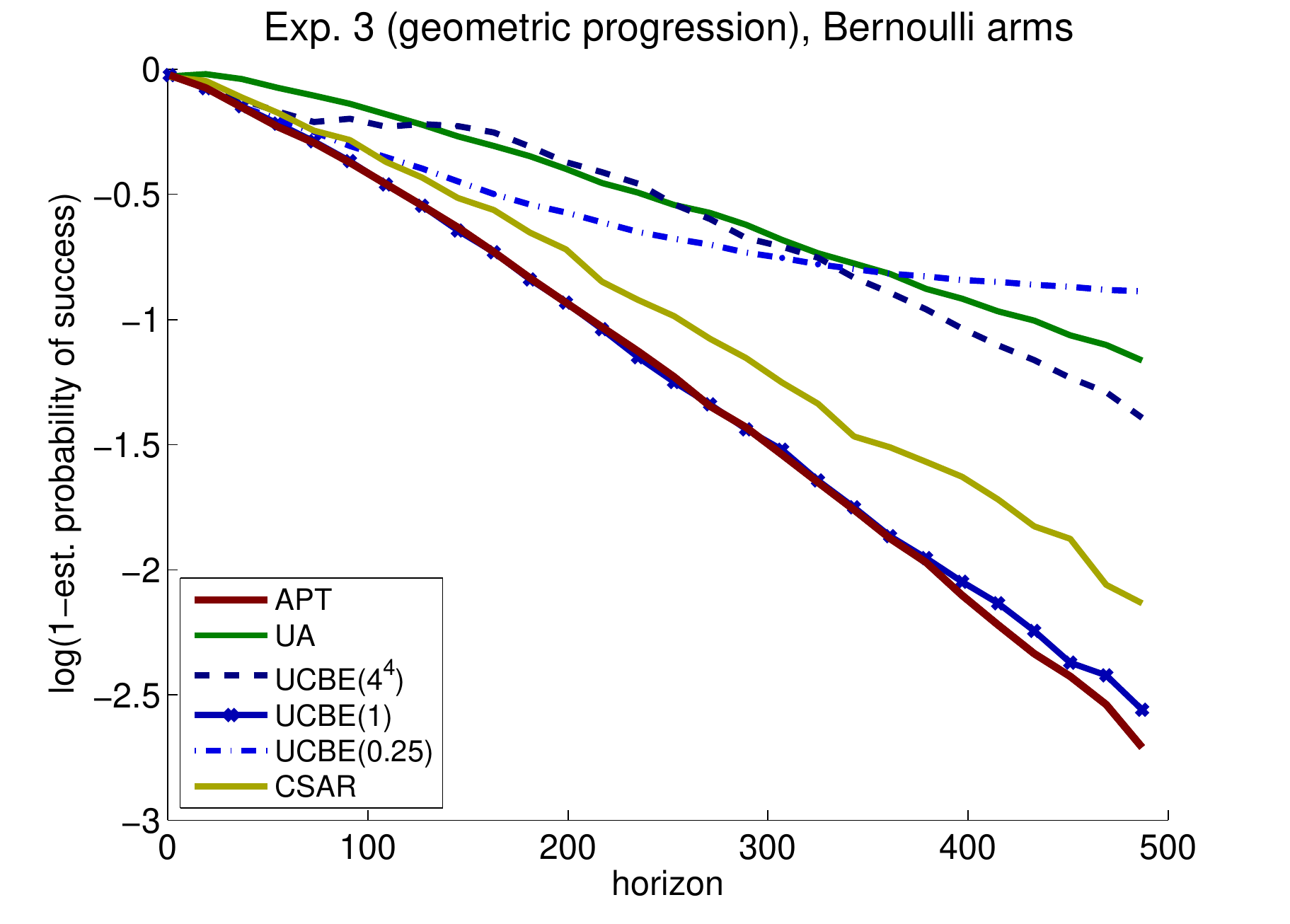}
\caption{Results of Experiments 1-3 with Bernoulli distributions. The average error of the specified methods is displayed on a logarithmic scale with respect to the horizon.}
\vspace{-0.5cm}
\end{figure*}

\paragraph{Remark} This implies that if $\mu^*$ is known to the learner, there exists an algorithm such that its probability of error is of order $\exp(-cT/H)$. The recent paper~\citep{carpentier2016} actually implies that the knowledge of $\mu^*$ is actually key here, since without this information, the simple regret is at least of order $\exp(-cT/(\log(K)H))$ in a minimax sense. 

\paragraph{Cumulative reward maximization}
In the cumulative reward maximization problem, the game setting is the same as the the one we considered but the aim of the learner is different : if we write $X_t$ for the sample collected at time $t$ by the algorithm, it aims at maximizing $\sum_{t \leq T} X_t$. The following proposition holds for our strategy APT that runs for $T$ times.
\begin{theorem}\label{thm:CR}
Let $K>0$, $R >0$ and $T \geq 2K$ and consider a problem where the distribution of the arms $\nu_k$ is $R$-sub-Gaussian.

Then APT run with parameters $\tau = \mu^*$ and $\epsilon = 0$ is such that
\vspace{-0.4cm}
\begin{align*}
T\mu^* - \mathbb E \sum_{t \leq T} X_t  &\leq \inf_{\delta \geq 1} \Big[\sum_{k \not = k^*} \frac{4R^2\log(T)\delta}{\mu^* - \mu_i}\\ 
&+ (\mu^*- \mu_i)(1 + \frac{K}{T^{2\delta - 2}})\Big].
\end{align*}
\end{theorem}
\vspace{-0.4cm}
This bound implies both the problem dependent upper bound of order $\sum_i \Delta_i^{-1}\log(T)$ and the problem independent upper bound of order $\sqrt{TK\log(T)}$, and this matches the performance of algorithms like UCB for any tuning parameter. A similar result can also be found in~\cite{salomon2011deviations}.

\paragraph{Discussion} Propositions~\ref{thm:Simple} and~\ref{thm:CR}, whose proofs are provided in Appendix \ref{app:A}, imply that our algorithm APT is a good strategy for solving \textit{at the same time} both problems when $\mu^*$ is known. As mentioned previously, this is counter intuitive since one would expect a good strategy for the best arm identification problem to explore significantly more than a good strategy for the cumulative reward maximization problem. To convince oneself, it is sufficient to look at the two-armed case, for which in the fixed budget it is optimal to sample both arms equally, while this strategy has linear regret in the cumulative setting. This intuition is formalized in \cite{bubeck2009pure} where the authors prove that no algorithm can achieve this without additional information. Our results therefore imply that the knowledge of $\mu^*$ by the learner is a sufficient information so that Theorem 1 of \cite{bubeck2009pure} does not hold anymore and there exists algorithms that solve both problems at the same time, as APT does.

\paragraph{TopM problem} An extension of the best arm identification problem is known as TopM arms identification problem, where one is concerned with identifying the set of the M arms with the highest means~\cite{bubeckmultiple,gabillon2012best, kaufmann2014complexity, zhou2014optimal,chen2014combinatorial, cao2015top}. If the learner has some additional information, such as the mean values of the arms with $M$th and $(M+1)$th highest means, then it is straightforward that one can apply our algorithm APT, setting $\tau$ in the middle between the $M$th and $(\textsc{M}+1)$th highest means. The set $\widehat{S}_{\tau}$ would then be returned as the estimated set of $M$ optimal arms. The upper bound and proof for this problem is a direct consequence of Theorem~\ref{thm:UB}, and granted one has such extra-information, outperforms existing results for the fixed budget setting, see~\cite{bubeckmultiple,kaufmann2014complexity,chen2014combinatorial, cao2015top}. If the complexity $H$ were also known to the learner, the strategy in~\cite{gabillon2012best} would attain a similar performance.

\section{Experiments}\label{experiments}


We illustrate the performance of algorithm APT in a number of experiments. For comparison, we use the following methods which include the state of the art CSAR algorithm of~\cite{chen2014combinatorial} and two minor adaptations of known methods that are also suitable for our problem.\\
\textbf{Uniform Allocation (UA):} For each $t\in\{1,2,\ldots,T\}$, we choose $I_t\sim\mathcal{U}_{\mathbb{A}}$. This method is known to be optimal if all arms are equally difficult to classify, that is in our setting, if the quantities $\Delta_i^{\tau,\epsilon}$, $i\in\mathbb{A}$, are very close.\\
\textbf{UCB-type algorithm:} The algorithm UCBE given and analyzed in \cite{audibert2010best} is designed for finding the best arm - and its heuristic is to pull the arm that maximizes a UCB bound - see also~\cite{gabillon2012best} for an adaptation of this algorithm to the general TopM problem.  
The natural adaptation of the method for our problem corresponds to pulling the arm that minimizes $\widehat{\Delta}_k(t)-\sqrt{\frac{a}{T_k(t)}}$. From the theoretical analysis in the paper~\cite{audibert2010best, gabillon2012best}, it is not hard to see that setting $a \approx (T-K)/H$ minimizes their upper bound, and that this algorithm attains the same expected loss as ours - but it requires the knowledge of $H$. In the experiments we choose values $a_i=4^{i}\frac{T-K}{H}$, $i\in\{-1,0,4\}$, and denote the respective results as UCBE$(4^i)$. The value $a_0$ can be seen as the optimal choice, while the two other choices give rise to strategies that are sub-optimal because they respectively explore too little or too much.\\
\textbf{CSAR:} As mentioned before, this method is given in \cite{chen2014combinatorial}. In our specific setting, via the shift $\widetilde{\mu}_i=\mu_i-\tau$, the lines 7-17 of the algorithm reduce to classifying the arm $i$ that maximizes $|\widetilde{\mu}_i|$ based on its current mean. The set $A_t$ corresponds to $\widehat{S}_\tau$ at time $t$. In fact in our specific setting the CSAR algorithm is a successive reject-type strategy (see~\cite{audibert2010best} where the arm whose empirical mean is furthest from the threshold is rejected at the end of each phase.\\

Figure 1 displays the estimated probability of success on a logarithmic scale with respect to the horizon of the six algorithms based on $N=5000$ simulated games with $\tau=\frac{1}{2}$, $\epsilon=0.1$, $K=10$, and $T=500$.\\
\textbf{Experiment 1 (3 groups setting):} $K$ Bernoulli arms with means $\mu_{1:3}\equiv 0.1,$ $\mu_{4:7}=(0.35,0.45,0.55,0.65)$ and $\mu_{8:10}\equiv 0.9$, which amounts to 2 difficult relevant arms (that is, outside the $2\epsilon$- band), 2 difficult irrelevant arms and six easy relevant arms.\\
\textbf{Experiment 2 (arithmetic progression):} $K$ Bernoulli arms with means $\mu_{1:4}=0.2+(0:3)\cdot 0.05,$ $\mu_5=0.45$, $\mu_6=0.55$ and $\mu_{7:10}=0.65+(0:3)\cdot 0.05$, which amounts to 2 difficult irrelevant arms and eight arms arithmetically progressing away from $\tau$.\\
\textbf{Experiment 3 (geometric progression):} $K$ Bernoulli arms with means $\mu_{1:4}=0.4-0.2^{1:4}$, $\mu_5=0.45$, $\mu_6=0.55$ and $\mu_{7:10}=0.6+d^{5-(1:4)}$, which amounts to 2 difficult irrelevant arms and eight arms geometrically progressing away from $\tau$.

The experimental results confirm that our algorithm may only be outperformed by methods that have an advantage in the sense that they have access to the underlying problem complexity and, in the case of UCBE$(1)$, an additional optimal parameter choice. In particular, other choices for that parameter lead to significantly less accurate results comparable to the naive strategy UA. These effects are also visible in the further results given in Appendix \ref{AppExp}. 

\paragraph{Conclusion} In this paper we proposed a parameter free algorithm based on a new heuristic for the TBP problem in the fixed confidence setting - and we prove that it is optimal which is a kind of result which is highly non trivial for combinatorial pure exploration problems with fixed budget.

\paragraph{Acknowledgement} This work is supported by the DFG's Emmy Noether grant MuSyAD (CA 1488/1-1).

\bibliography{library}
\bibliographystyle{icml2016}

\clearpage
\begin{appendix}
\onecolumn
\section{Proofs}\label{app:A}
\subsection{Proof of Theorem \ref{thm:LB}}
\begin{proof}
In this proof, we will prove that on at least one instance of the problem, any algorithm makes a mistake of order at least $\exp(-cT/H)$.

{\bf Step 0: Setting and notations.} Let us consider $K$ real numbers $\Delta_i \geq 0$, and let us set $\tau=0, \epsilon = 0$. Let us write $\nu_i:=\mathcal N(\Delta_i,1)$ for the Gaussian distribution of mean $\Delta_i$ and variance $1$, and $\nu_i':=\mathcal N(-\Delta_i,1)$ for the Gaussian distribution of mean $-\Delta_i$ and variance $1$.
Note that this construction is easily generalised to cases where $\tau\neq 0$ or $\epsilon\neq 0$ by translation or careful choice of the $\Delta_i$. 

We define the product distributions $\mathcal{B}^i$ where $i \in \{0, ..., K\}$ as $\nu_1^i \otimes ... \otimes \nu_K^i$ where for $k \leq K$, $\nu_k^i := \nu_i \mathbf 1_{k \neq i} + \nu_i' \mathbf 1_{k = i}$ is $\nu_i$ if $k \neq i$ and $\nu_i'$ otherwise. We also extend this notation to $\mathcal{B}^0$, where none of the arms is flipped with respect to the threshold ($\forall k$, $\nu_k^0:= \nu_i$).  It is straightforward that the gap $\Delta_i$ of arm $i$ with respect to the threshold $\tau = 0$ does not depend on $\mathcal{B}^i$ and is equal to $\Delta_i$. It follows that all these problems have the same complexity $H$ as defined previously (with $\epsilon = 0$ and $\tau = 0$). 

We write for $i \leq K$, $\mathbb P_{\mathcal B}^i$ for the probability distribution according to all the samples that a strategy could possibly collect up to horizon $T$, i.e.~according to the samples $(X_{k,s})_{k \leq K, s \leq T} \sim (\mathcal{B}^i)^{\otimes T}$. Let $(T_k)_{k \leq K}$ denote the numbers of samples collected by the algorithm on arm $k$.


Let $k \in \{0, ..., K\}$. Note that
$$ \text{KL}_k := \text{KL}(\nu_k', \nu_k) =  2\Delta_k^2,$$
where $\text{KL}$ is the Kullback Leibler divergence. Let $T \geq t \geq 0$. We define the quantity:
\begin{align*}
\widehat{\text{KL}}_{k,t} &= \frac{1}{t} \sum_{s=1}^t \log(\frac{d \nu_k'}{d \nu_k}(X_{k,s}))= -\frac{1}{t} \sum_{s=1}^t 2X_{k,s} \Delta_k.
\end{align*}


{\bf Step 1: Concentration of the empirical KL.} Let us define the event:
\begin{align*}
\xi &= \Big\{\forall k\leq K, \forall t\leq T, |\widehat{\text{KL}}_{k,t} - \text{KL}_k|\leq 4\Delta_k \ \sqrt[]{\frac{\log(4(\log(T)+1)K)}{t}} \Big\}.
\end{align*}
Since $\widehat{\text{KL}}_{k,t} = -\frac{1}{t} \sum_{s=1}^t 2X_{k,s} \Delta_k$ and $\text{KL}_k = 2\Delta_k^2$, by Gaussian concentration (a peeling and the maximal martingale inequality), it holds that for any $i$ that $\mathbb P_{\mathcal B^i}(\xi) \geq 3/4$.



{\bf Step 2: A change of measure.} We will now use the change of measure introduced previously for a well chosen event $\mathcal{A}$. Namely, we consider $\mathcal{A}_i = \{ i \in \widehat{S}_\tau\}$, the event where the algorithm classified arm $i$ as being above the threshold. We have by doing a change of measure between $\mathcal{B}^i$ and $\mathcal{B}^0$ (since they only differ in arm $i$ and only the $T_i$ first samples of arm $i$ by the algorithm):
\begin{align*}
\mathbb{P}_{\mathcal{B}^i}(\mathcal{A}_i) & =\mathbb{E}_{\mathcal{B}^0}\Big[\mathbf{1}_{\mathcal{A}_i}\exp\big(-T_i\widehat{\text{KL}}_{i,T_i}\big)\Big]\\
& \geq \mathbb{E}_{\mathcal{B}^0}\Big[\mathbf{1}_{\mathcal{A}_i\cap \xi}\exp\big(-T_i\widehat{\text{KL}}_{i,T_i}\big)\Big]\\
& \geq \mathbb{E}_{\mathcal{B}^0}\Big[\mathbf{1}_{\mathcal{A}_i\cap \xi}\exp\big(-2\Delta_i^2 T_i -4\Delta_i\sqrt{T_i} \ \sqrt[]{\log((4\log(T)+1)K)}\big)\Big],
\end{align*}
by definition of $\xi$ and $\text{KL}_i$.
\\
\\
{\bf Step 3: A union of events.} We now consider the event $\mathcal{A} = \bigcap\limits_{i=1}^{K} \mathcal{A}_i$, i.e.~the event where all arms are classified as being above the threshold $\tau = 0$. We have:
\begin{flalign}
\max\limits_{i \in \{1,\ldots,K\}}  \mathbb{P}_{\mathcal{B}^i}(\mathcal{A}_i)
 & \geq \frac{1}{K} \sum\limits_{i=1}^K \mathbb{P}_{\mathcal{B}^i}(\mathcal{A}_i)\\
 &\geq \frac{1}{K} \sum\limits_{i=1}^K \mathbb{P}_{\mathcal{B}^i}(\mathcal{A}_i\cap \xi)\nonumber\\
&\geq \frac{1}{K}  \sum\limits_{i=1}^K \mathbb{E}_{\mathcal{B}^0}\Big[ \mathbf{1}_{\mathcal{A}_i \cap \xi} \exp\big(-2T_i \Delta_i^2 -4\Delta_i\sqrt{T_i} \ \sqrt[]{\log(4(\log(T)+1)K)}\big)\Big]\nonumber\\
&\geq  \mathbb{E}_{\mathcal{B}^0}\Big[\mathbf{1}_{\mathcal{A} \cap \xi}  \frac{1}{K} \sum\limits_{i=1}^K \exp\big(-3T_i \Delta_i^2 -4\log(4(\log(T)+1)K) \big)\Big]\nonumber\\
&\geq  \exp\big(-4\log(4(\log(T)+1)K)\big)  \mathbb{E}_{\mathcal{B}^0}\Big[\mathbf{1}_{\mathcal{A} \cap \xi}  S\Big],\label{eq:LBsum}
\end{flalign}
where the fourth line comes from using $2ab \leq a^2 + b^2$ with $a = \Delta_i \sqrt{T_i}$ and where:
$$S = \dfrac{1}{K} \sum\limits_{i=1}^K \exp\big(-3T_i\Delta_i^2 \big).$$
Since $\sum_i T_i = T$ and all $T_i$ are positive, there exists an arm $i$ such that $T_i \leq \frac{T}{H\Delta_i^2}$. This yields:
$$S \geq \frac{1}{K}\exp\big(-\frac{3T}{H}\big) = \exp\big(-\frac{3T}{H} - \log(K)\big).$$
This implies by definition of the risk:
\begin{align*}
\max\limits_{i \in \{0,\ldots,K\}}  \mathbb{E}_{\mathcal{B}^i}(\mathcal L(T)) & \geq \max\Big(\max\limits_{i \in \{1,\ldots,K\}}  \mathbb{P}_{\mathcal{B}^i}(\mathcal{A}_i), 1-\mathbb{P}_{\mathcal{B}^0}(\mathcal{A})\Big)\\
&\geq  \frac{1}{2}\exp\big(-\frac{3T}{H}-4\log(4(\log(T)+1)K)\big)- \log(K) \mathbb{E}_{\mathcal{B}^0}\Big[\mathbf{1}_{\mathcal{A} \cap \xi} \Big] +\frac{1}{2} (1-\mathbb{P}_{\mathcal{B}^0}(\mathcal{A}))\\
&=\frac{1}{2}  \exp\big(-\frac{3T}{H}-4\log(4(\log(T)+1)K-\log(K))\big)  \mathbb{P}_{\mathcal{B}^0}\Big[\mathcal{A} \cap \xi \Big]+\frac{1}{2} (1-\mathbb{P}_{\mathcal{B}^0}(\mathcal{A}))\\
&\geq \frac{1}{8}\exp\big(-\frac{3T}{H}-4\log(4(\log(T)+1)K)-\log(K)\big)\\
&\geq \exp\big(-\frac{3T}{H}-4\log(12(\log(T)+1)K)\big),
\end{align*}
The fourth line comes from $\mathbb{P}(\xi) \geq 3/4$, and we consider two cases $\mathbb{P}_{\mathbb{B}^0}(\mathcal{A}) \geq 1/2$ and $\mathbb{P}_{\mathbb{B}^0}(\mathcal{A}) \leq 1/2$. The first leads directly to the condition as the intersection is at least of probability $1/4$;
in the latter case, we have the same bound via
$$\max\limits_{i \in \{0,\ldots,K\}}  \mathbb{E}_{\mathcal{B}^i}(\mathcal L(T))\geq \mathbb{E}_{\mathcal{B}^0}(\mathcal L(T))=\mathbb{P}_{\mathcal{B}^0}(\mathcal{A}^C)\geq 1/2.$$
This concludes the proof.

\end{proof}

\subsection{Proof of Theorem \ref{thm:UB}}
\begin{proof} In this proof, we will show that on a well chosen event $\xi$, we classify correctly the arms which are over $\tau + \epsilon$, and reject the arms that are under $\tau - \epsilon$.\\

{\bf Step 1: A favorable event.} Let $\delta = (4\sqrt{2})^{-1}$. Towards this goal, we define the event $\xi$ as follows:
$$\xi = \Big\{\forall i \in \mathbb{A}, \forall s \in \{1, ..., T\}:  |\frac{1}{s}\sum_{t=1}^sX_{i,t} - \mu_i| \leq \sqrt{\frac{T\delta^2}{Hs}}\Big\}.$$

We know from Sub-Gaussian martingale inequality that for each $i \in \mathbb{A}$ and each $u \in \{0, ..., \lfloor\log(T)\rfloor\}$:

\begin{align*}
\mathbb{P}\Big(\exists v \in[2^u,2^{u+1}], \{  |\frac{1}{v}\sum_{t=1}^vX_{i,t} - \mu_i| \geq \sqrt{\frac{T\delta^2}{Hv}} \}\Big) \leq \exp(-\frac{T\delta^2}{2R^2H}).
\end{align*}

$\xi$ is the union of these events for all $i \leq K$ and $s\leq  \lfloor\log(T)\rfloor$. As there are less than $(\log(T)+1)K$ such combinations, we can lower-bound its probability of occurrence with a union bound by:

$$\mathbb{P}(\xi) \geq 1 - 2(\log(T)+1)K\exp(-\frac{T\delta^2}{2R^2H}).$$

{\bf Step 2: Characterization of some helpful arm.} At time $T$, we consider an arm $k$ that has been pulled after the initialization phase and such that $T_k(T)-1 \geq \frac{(T-K)}{H\Delta_k^2}$. We know that such an arm exists otherwise we get:
$$T-K = \sum_{i = 1}^K (T_i(T)-1) < \sum_{i = 1}^K \frac{T-K}{H\Delta_i^2} = T-K,$$
which is a contradiction. Note that since $T \geq 2K$, we have that $T_k(T)-1 \geq \frac{T}{2H\Delta_k^2}$\\
We now consider $t \leq T$ the last time that this arm $k$ was pulled. Using $T_k(t)\geq 2$ (by the initialisation of the algorithm), we know that:
\begin{equation}\label{eq:Ti}
T_k(t) \geq T_k(T) - 1  \geq  \frac{T}{2H\Delta_k^2}.
\end{equation}
{\bf Step 3: Lower bound on the number of pulls of the other arms.} On $\xi$, at time $t$ as we defined previously, we have for every arm $i$:
\begin{equation}\label{hoeffding}
|\hat{\mu}_i(t) - \mu_i| \leq \sqrt[]{\frac{T\delta^2}{HT_i(t)}}.
\end{equation}

From the reverse triangle inequality and Equation~\eqref{gap_k_t}, we have:
\begin{align*}\label{triangle}
|\hat{\mu}_i(t) - \mu_i| &= |(\hat{\mu}_i(t) - \tau) - (\mu_i - \tau)|\\
& \geq  ||\hat{\mu}_i(t) - \tau| - |\mu_i - \tau||\\
& \geq  |(|\hat{\mu}_i(t) - \tau| + \epsilon) - (|\mu_i - \tau| + \epsilon)|\\
&\geq |\widehat{\Delta}_i(t) - \Delta_i|.
\end{align*}

Combining this with \eqref{hoeffding} yields the following:
\begin{equation}\label{bounding_delta}
\Delta_k - \sqrt[]{\frac{T\delta^2}{HT_k(t)}} \leq \widehat{\Delta}_k(t) \leq \Delta_k + \sqrt[]{\frac{T\delta^2}{HT_k(t)}}.
\end{equation}

By construction, we know that at time $t$ we pulled arm $k$, which yields for every $i \in \mathbb{A}$:
\begin{equation}\label{algo_cond}
B_k(t) \leq B_i(t).
\end{equation}

We can lower bound the left-hand side of \eqref{algo_cond} using \eqref{eq:Ti}:
\begin{align}\label{eq:LB}
\Big(\Delta_k - \sqrt[]{\frac{T\delta^2}{HT_k(t)}}\Big)\sqrt{T_k(t)} & \leq B_k(t) \nonumber \\
\Big(\Delta_k - \sqrt{2}\delta \Delta_k\Big)\sqrt{\frac{T}{2H\Delta_k^2}} & \leq B_k(t) \nonumber \\ 
\Big(\frac{1}{\sqrt{2}} - \delta\Big)\sqrt{\frac{T}{H}} & \leq B_k(t),
\end{align}
and upper bound the right hand side using \eqref{bounding_delta} by:
\begin{align}\label{eq:UB}
B_i(t) &= \widehat{\Delta}_i\sqrt{T_i(t)} \nonumber \\
& \leq \Big(\Delta_i + \sqrt{\frac{T\delta^2}{HT_i(t)}}\Big)\sqrt{T_i(t)} \nonumber \\
& \leq \Delta_i\sqrt{T_i(t)} + \delta\sqrt{\frac{T}{H}}.
\end{align}

As both $\widehat{\Delta}_i$ and $\Delta_i$ are positive by definition, combining \eqref{eq:LB} and \eqref{eq:UB} yields the following lower bound on $T_i(T) \geq T_i(t)$:
\begin{equation}\label{lb_Ti}
\Big(1- 2\sqrt{2}\delta\Big)^2\frac{T}{2H\Delta_i^2} \leq T_i(T).
\end{equation}

{\bf Step 4: Conclusion.} On $\xi$, as $\Delta_i$ is a positive quantity, combining \eqref{hoeffding} and \eqref{lb_Ti} yields:
\begin{equation}\label{eq:mu_i}
\mu_i - \Delta_i \frac{\sqrt{2}\delta}{1-2\sqrt{2}\delta} \leq \hat{\mu}_i(T) \leq \mu_i + \Delta_i \frac{\sqrt{2}\delta}{1-2\sqrt{2}\delta},
\end{equation}
where $\frac{\sqrt{2}\delta}{1-2\sqrt{2}\delta}$ simplifies to $1/2$ for $\delta = (4\sqrt{2})^{-1}$.\\
For arms such that $\mu_i \geq \tau +\epsilon$, then $\Delta_i = \mu_i - \tau + \epsilon$ and we can rewrite \eqref{eq:mu_i}:
\begin{align*}
\mu_i - \tau -  \frac{1}{2}\Delta_i &\leq \hat{\mu}_i(T) - \tau\\
(\mu_i - \tau)(1 - \frac{1}{2}) - \frac{\epsilon}{2} &\leq \hat{\mu}_i(T) - \tau\\
0 & \leq \hat{\mu}_i(T) - \tau,
\end{align*}
where the last line uses $\mu_i \geq \tau +\epsilon$. One can easily check through similar derivations that $\hat{\mu}_i(T) - \tau < 0$ holds for $\mu_i < \tau - \epsilon$. On $\xi$, arms over $\tau + \epsilon$ are all accepted, and arms under $\tau - \epsilon$ are all rejected, which means the loss suffered by the algorithm is $0$. As $1 - \mathbb{P}(\xi) \leq 2(\log(T)+1)K\exp(-\frac{1}{64R^2}\frac{T}{H})$, this concludes the proof.
\end{proof}

\subsection{Proof of Theorem \ref{thm:Simple}}
\begin{proof}
We will prove that on a well defined event $\xi$, sub-optimal arms are pulled at most $\frac{T}{2\Delta_k^2H} - 1$ times, which translates to the best arm being chosen at the end of the horizon as it was pulled more than half of the time.

{\bf Step 1: A favorable event.} Let $\delta = 1/18$. We define the following events $\forall i \in \mathbb{A}$:
\begin{equation}
\xi_{i} = \{ \forall s \leq T : |\mu^* - \widehat{\mu}_{i}(s)| \leq \sqrt{\frac{T\delta}{HT_{i}(s)}} \}, \nonumber
\end{equation}
We now define $\xi$ as the intersection of these events:
\begin{equation}
\xi~=~\bigcap\limits_{k\in\mathbb{A}}~\xi_k.\nonumber
\end{equation}
Using the same Sub-Gaussian martingale inequality as in the proof of Theorem \ref{thm:UB}, we can lower bound its probability of occurrence with a union bound by:
\begin{equation}
P(\xi) \geq 1 - 2(\log(T)+1)K\exp(-\frac{T}{36R^2H}) \nonumber
\end{equation}

{\bf Step 2: The wrong arm at the wrong time.} Let us now suppose that a sub-optimal arm $k$ was pulled at least $\frac{T-K}{2\Delta_k^2H}$ times after the initialization which translates to $T_k(T) - 1 \geq \frac{T-K}{2\Delta_k^2H}$. Let us now consider the last time $t \leq T$ that this arm was pulled. As it was pulled at time $t$, the following inequality holds:
\begin{equation}\label{algo2}
B_k(t) \leq B_{k^*}(t).
\end{equation}
On $\xi$, we can now lower bound the left hand side by:
\begin{align}\label{lb2}
(\Delta_k - \sqrt{\frac{T\delta}{HT_k(t)}})\sqrt{T_k(t)} &\leq B_{k}(t)\nonumber\\
\Delta_k\sqrt{T_k(t)} - \sqrt{\frac{T\delta}{H}} &\leq B_{k}(t),
\end{align}
We also upper bound the right hand side of \eqref{algo2} by:
\begin{align} \label{ub2}
B_{k^*}(t) & \leq \sqrt{\frac{T\delta}{H}}.
\end{align}
Combining both bounds \eqref{lb2} and \eqref{ub2} with \eqref{algo2}, as well as rearranging the terms yields:
\begin{align}\label{eq:contr}
\Delta_k \sqrt{T_k(t)} & \leq 2\sqrt{\frac{T\delta}{H}} \nonumber \\
T_k(t)\Delta_k^2 & \leq \frac{4T\delta}{H}.
\end{align}
Using $T_k(t) \geq T_k(T) - 1 \geq \frac{T-K}{2\Delta_k^2H}$ as well as $T \geq 2K$, we have
\begin{equation}
T_k(t) \geq \frac{T}{4\Delta_k^2H}.
\end{equation}
Plugging this in \eqref{eq:contr} brings the following condition:

\begin{equation}
\frac{T}{4\Delta_k^2H}\Delta_k^2 \leq \frac{4T\delta}{H}.
\end{equation}

which directly reduces to $\delta \geq 1/16$, which is a contradiction as we have set $\delta = 1/18$.

As we have proved that for any sub-optimal arm $i \not = k^*$ it satisfies $T_i(T) < \frac{T}{2\Delta_i^2H}$, summing for all arms yields:
\begin{align}
T-T_{k^*}(T) & = \sum_{i \not = k^*} T_i(T) \nonumber\\
& < \frac{T}{2H} \sum_{i \not = k^*} \frac{1}{\Delta_i^2} = \frac{T}{2}.
\end{align}

We conclude by observing that $T_{k^*}(T) > T/2$, and as such will be chosen by the algorithm at the end as being the best arm.
\end{proof}

\subsection{Proof of Theorem \ref{thm:CR}}
\begin{proof}
In this proof we will show that with high probability the sub-optimal arms have been pulled at most at a logarithmic rate, and will then bound the expectation of the number of pulls of these arms.

{\bf Step 1: A favorable event.} We define the following events $\forall s \leq T$ :

\begin{equation}
\xi_{k^*,s} = \{\mu^* - \hat{\mu}_{k^*}(s) \leq R\sqrt{\frac{\log(T)\delta}{T_{k^*}(s)}} \}, \nonumber
\end{equation}
as well as for all arms $i \not = k^*$:
\begin{equation}
\xi_{i,s} = \{\hat{\mu}_k(s) - \mu_k \leq R\sqrt{\frac{\log(T)\delta}{T_{k^i}(s)}} \}.\nonumber
\end{equation}

By Hoeffding's inequality, the complimentary $\bar{\xi}_k$ of each of these events has probability at most $T^{-2\delta}$.\\
We now consider $\xi$ the intersection of these events for all $k \in \mathbb{A}$. By a union bound, as there are $T$ such events for each arm, we have:
\begin{equation}
\mathbb{P}(\xi) \geq 1 - \frac{K}{T^{2\delta-1}}.
\end{equation}

We also have:
\begin{equation}
\mathbb{P}(\bar{\xi}) \leq \frac{K}{T^{2\delta-1}}.
\end{equation}

We will now prove a bound on the number of pulls on $\xi$.

{\bf Step 2: Bound on pulls of sub-optimal arms.} We now consider the last time $t$ that arm $k \not = k^*$ was pulled, under the assumption that it was pulled at least once after the initialization. The decision rule of the algorithm yields:

\begin{equation}
B_k(t) \leq B_{k^*}(t).
\end{equation}

On $\xi$, we can now lower-bound the left-side and upper-bound the right hand side, which yields:
\begin{equation}
(\Delta_k - R\sqrt{\frac{\log(T)\delta}{T_k(t)}})\sqrt{T_k(t)} \leq R\sqrt{\frac{\log(T)\delta}{T_{k^*}(t)}}\sqrt{T_{k^*}(t)},
\end{equation}
which can be rearranged as such:
\begin{equation}
\Delta_k\sqrt{T_k(t)} \leq 2R\sqrt{\log(T)\delta},
\end{equation}
and the following bound on $T_k(T)$:

\begin{equation}
T_k(T) \leq \frac{4R^2\log(T)\delta}{\Delta_k^2}+1.
\end{equation}

Note that we here make the assumption that the arm was pulled at least once by the algorithm after the initialization. If it has only been pulled during the initialization, the bound still trivially holds as we have at least one pull.

{\bf Step 3: Conclusion.} We can thus upper-bound the expectation of $T_k(t)$, as when $\xi$ does not hold we get at most $T$ pulls:
\begin{equation}
\mathbb{E}[T_k(T)] \leq \frac{4R^2\log(T)\delta}{\Delta_k^2} + 1 + \frac{K}{T^{2\delta - 2}},
\end{equation}
and we get the following bound on the pseudo-regret when $\xi$ holds:
\begin{equation}
\bar{R}_T \leq \sum_{k \not = k^*} \frac{4R^2\log(T)\delta}{\Delta_k} + \Delta_k(1 + \frac{K}{T^{2\delta - 2}}).
\end{equation}

Plugging $\delta = 1$ yields:

\begin{equation}
\bar{R}_T \leq \sum_{k \not = k^*} \frac{4R^2\log(T)}{\Delta_k} + \Delta_k(1 + K),
\end{equation}
and we recover the classical bound of the UCB1 algorithm.
\end{proof}
\newpage
\section{Further Experimental Results}\label{AppExp}
We now also provide simulation results for our three settings in the case of Gaussian arms with means $\mu_i$ and variances $\sigma_i^2=0.25$. Again, only the correctly tuned UCBE- algorithm outperforms APT.
\begin{figure}[h]
\centerline{\includegraphics[scale=0.3]{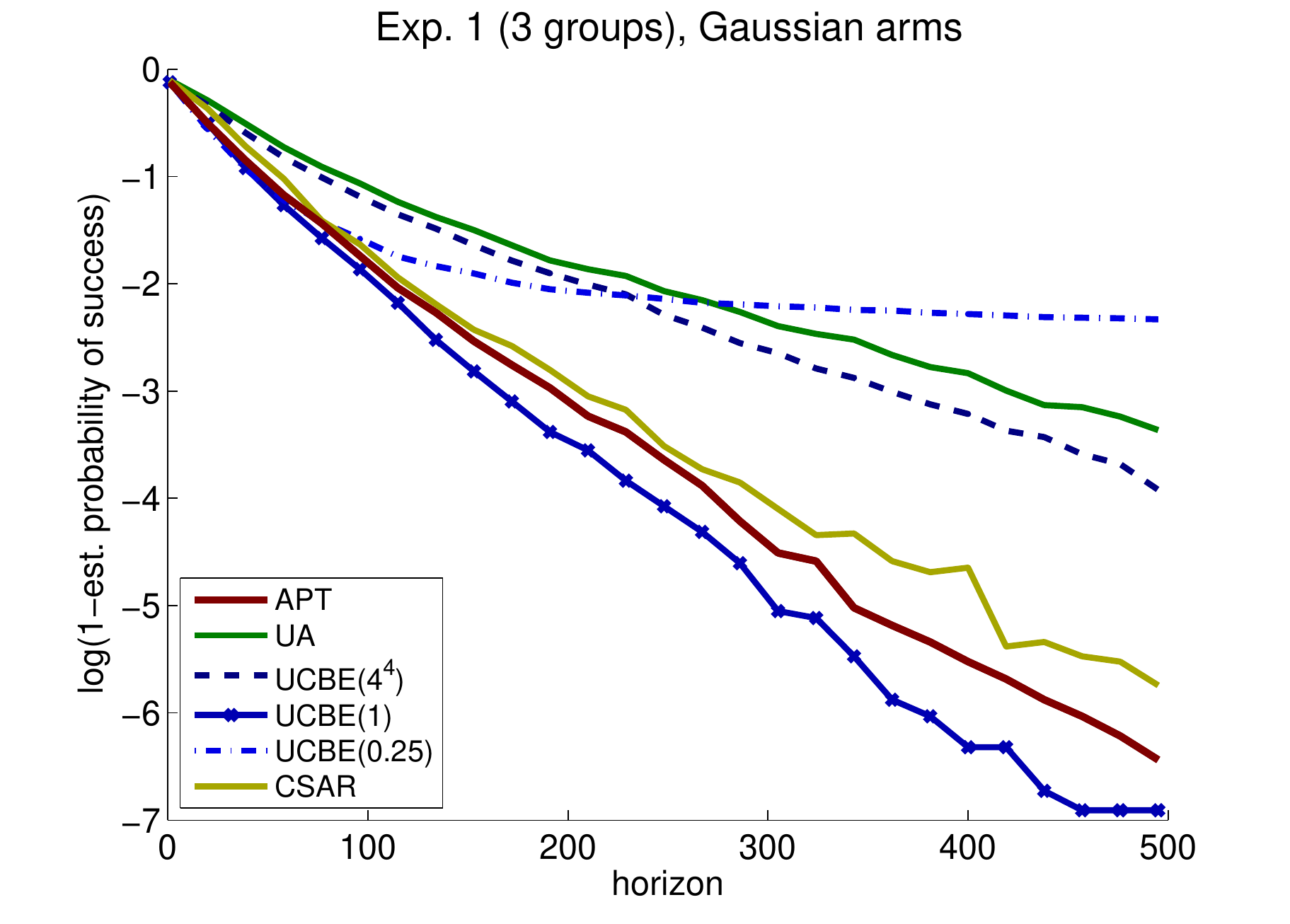}\  \includegraphics[scale=0.3]{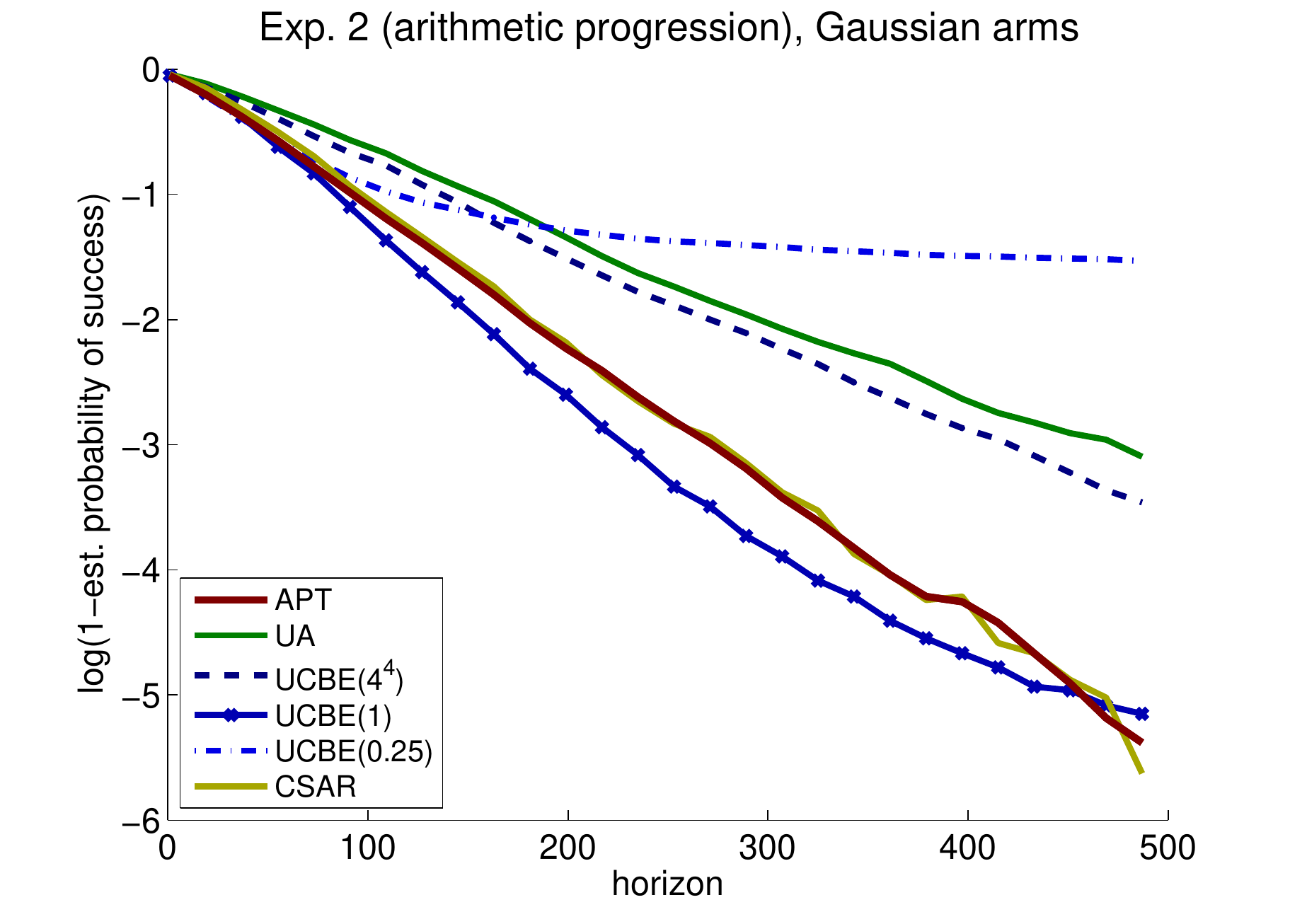}\ 
\includegraphics[scale=0.3]{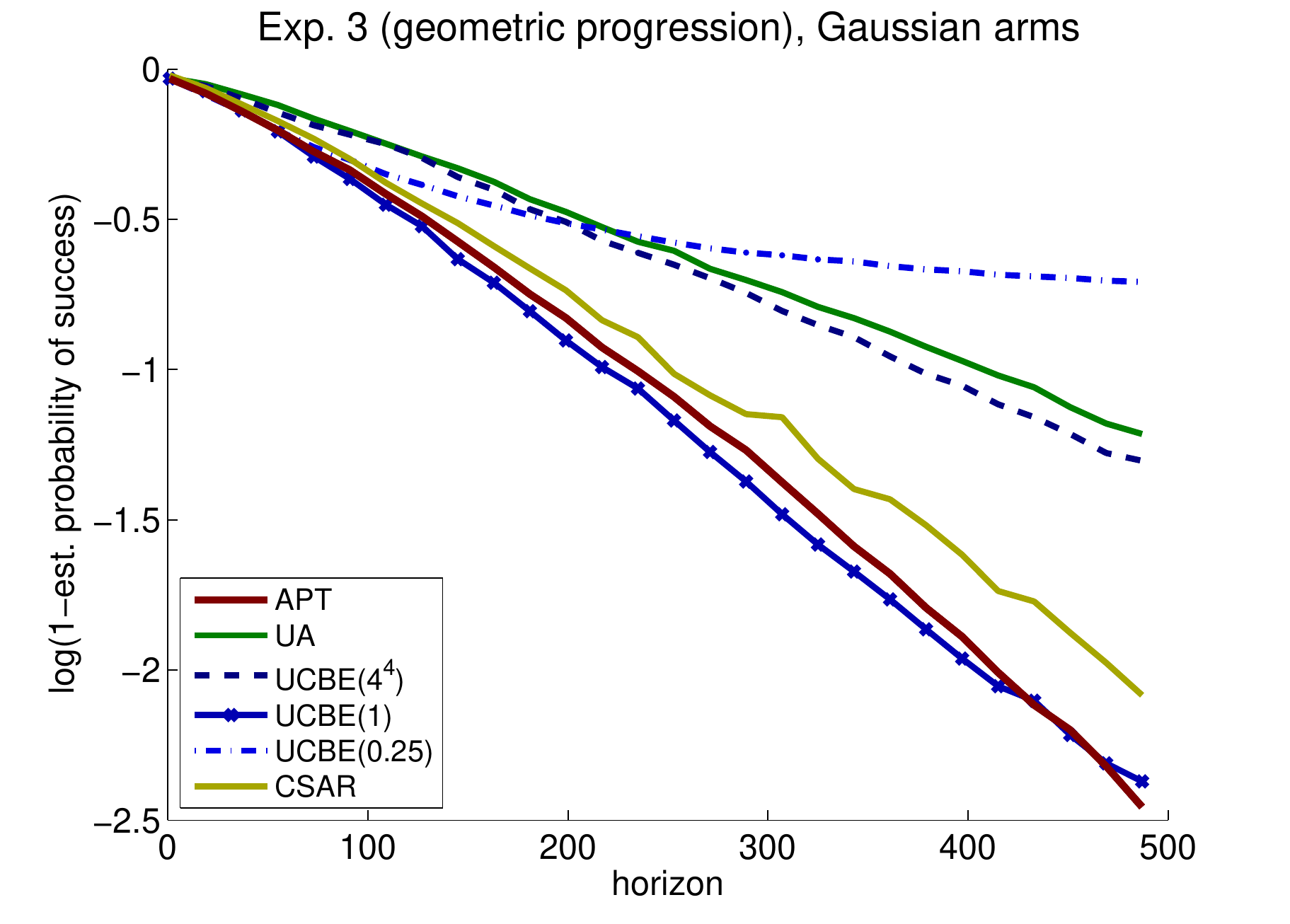}}
\caption{Results of Experiments 1-3 with Gaussian distributions. The average error of the specified methods is displayed on a logarithmic scale with respect to the horizon.}
\end{figure}
\end{appendix}
\end{document}